\documentclass{article} %
\usepackage{colm2024_conference}

\usepackage{microtype}
\usepackage{hyperref}
\usepackage{url}
\usepackage{booktabs}
\usepackage{graphicx}
\definecolor{darkblue}{rgb}{0, 0, 0.5}
\definecolor{forestgreen(web)}{rgb}{0.13, 0.55, 0.13}
\hypersetup{colorlinks=true, citecolor=darkblue, linkcolor=darkblue, urlcolor=darkblue}
\usepackage{siunitx}
\usepackage{amsmath}
\usepackage{bbding}
\usepackage{makecell}
\usepackage{wrapfig}

\title{Can Language Models Solve Olympiad Programming?}

\author{Quan Shi$^*$, Michael Tang$^*$, Karthik Narasimhan, Shunyu Yao  \\
Princeton Language and Intelligence (PLI), Princeton University\\
\texttt{\{qbshi, mwtang\}@princeton.edu} \\
}

\newcommand{\greencheck}{\color{forestgreen(web)}\Checkmark}
\newcommand{\redx}{\color{red}\XSolidBrush}

\begin{document}

\maketitle

\begin{abstract}
Computing olympiads contain some of the most challenging problems for humans, requiring complex algorithmic reasoning, puzzle solving, in addition to generating efficient code. However, it has been understudied as a domain to evaluate language models (LMs). In this paper, we introduce the USACO benchmark with 307 problems from the USA Computing Olympiad, along with high-quality unit tests, reference code, and official analyses for each problem. These resources enable us to construct and test a range of LM inference methods for competitive programming for the first time. We find GPT-4 only achieves a 8.7\% pass@1 accuracy with zero-shot chain-of-thought prompting, and our best inference method improves it to 20.2\% using a combination of self-reflection and retrieval over episodic knowledge. However, this is far from solving the benchmark. To better understand the remaining challenges, we design a novel human-in-the-loop study and surprisingly find that a small number of targeted hints enable GPT-4 to solve 13 out of 15 problems previously unsolvable by any model and method. Our benchmark, baseline methods, quantitative results, and qualitative analysis  serve as an initial step toward LMs with grounded, creative, and algorithmic reasoning.

\end{abstract}
\section{Introduction}
\def\thefootnote{*}\footnotetext{Equal contribution. Code, data, examples: \url{https://princeton-nlp.github.io/USACOBench/}.}\def\thefootnote{\arabic{footnote}}
Code generation has become an important domain to evaluate and deploy language models (LMs). However, with the scaling of LMs and the development of new inference methods~\citep{wei2022chain,shinn2023reflexion, chen2023teaching, zhou2022docprompting}, many popular coding benchmarks such as HumanEval~\citep{chen2021evaluating} and MBPP~\citep{austin2021program} have been saturated with solve rates above 90\%. 
To drive further progress, we need more challenging benchmarks that reveal limitations of existing models and inference methods, and provide actionable insights for improving LM's algorithmic reasoning.

Competitive programming is a natural fit for this pursuit, as it has been designed to rigorously evaluate the human ability to reason about complex scenarios and create novel algorithms. However, previous explorations of competitive programming lack exhaustive unit test suites, lack problem analyses, or lack enough problem diversity to comprehensively evaluate algorithmic reasoning \citep{li2022competition, hendrycks2021measuring, jain2024livecodebench}.

\begin{figure}[t]
  \centering
    \vspace{-35pt}
  \includegraphics[width=1\textwidth]{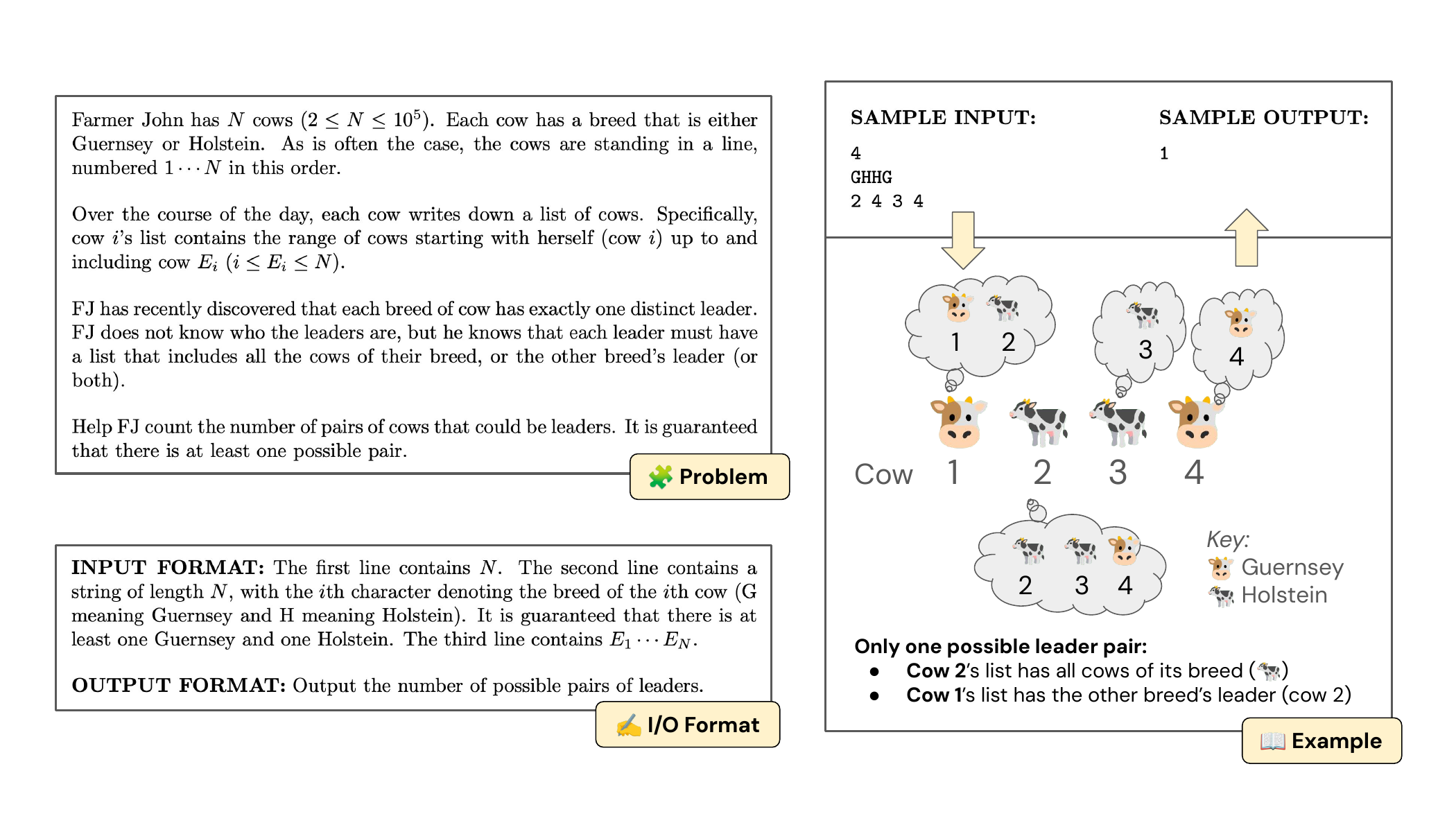} %
  \vspace{-20pt}
  \caption{Example USACO problem description, formatting instructions, and illustration (problem id: \href{https://usaco.org/index.php?page=viewproblem2&cpid=1275}{\texttt{1275\_bronze\_leaders}}). Solving this problem requires a combination of \textit{grounded} reasoning about the concept of leaders, \textit{creative} thinking to precisely count different cases of leader pairs, and \textit{algorithmic} reasoning to perform these ad hoc operations in linear time.}
  \label{sample_usaco_problem}
  \vspace{-10pt}
\end{figure}

We thus introduce \textbf{USACO}, a carefully crafted coding benchmark with 307 challenging problems from past USA Computing Olympiad (USACO) competitions. As shown in Figure~\ref{sample_usaco_problem}, each problem describes a task to solve in a fictional scenario, along with some example tuples of inputs, outputs, and explanations. Solving these problems not only require a wide range of algorithmic, mathematical, and commonsense knowledge, but also grounded and creative reasoning: unlike previous program synthesis benchmarks, successful models must reason over ad hoc environments, creating novel algorithms tailored to each problem scenario. On USACO, even the best LM (GPT-4) only reaches a zero-shot pass@1 solve rate of 8.7\% using zero-shot chain-of-thought prompting.

To study more advanced inference-time methods on competitive programming, for the first time, our benchmark also collects high-quality unit tests, reference code solutions, and official analysis for each problem, along with corresponding instructional texts in the form of competition programming textbooks. Based on these resources, we construct a range of baseline methods based on retrieval~\citep{gao2023retrieval}, self-reflection~\citep{shinn2023reflexion, chen2023teaching}, and their combinations. We find that combining retrieval over similar problems and solutions and self-reflection maximizes performance gains, well over \textbf{doubling} the zero-shot solve rate of GPT-4. However, all methods are still far from solving the benchmark above bronze level, the easiest difficulty tier.

To further understand the limitations and potentials of LM reasoning toward competitive programming, we perform a novel human study where humans interact with LMs in a conversational ``tutoring'' setup by pointing out errors and giving minimal hints. To our surprise, on a subset of 15 problems where GPT-3.5 and GPT-4 can never solve using any inference methods, such a human-in-the-loop setup leads to GPT-4 solving 13 out of 15 problems, whereas GPT-3.5 solves \textit{none}. This indicates the emergent potential of stronger LMs to incorporate high-quality feedback, the need to develop new methods that can generate such human-level corrective feedback, and a re-thinking of the right metric for measuring model capabilities beyond the overly strict execution success.

To summarize, our contributions of our work are:
\begin{itemize}
\item We propose the USACO benchmark, the first benchmark based on olympiad programming with high quality test cases, problem analyses, and auxiliary resources. 

\item For the first time, we construct and test LM inference methods for Olympiad programming, such as self-reflection and retrieval. Our results indicate a combination of retrieval and self-reflection can significant boost performance, but is still far from solving the benchmark.

\item We conduct a novel human-in-the-loop study to characterize the capabilities and limitations of LMs for Olympiad programming, complementary to automatic experiments based on execution success. We find that only certain models can successfully incorporate feedback, uncovering latent differences between models
\end{itemize}

\section{Related Work}
\paragraph{Code Generation Benchmarks} Language model performance on simple program synthesis has been thoroughly explored \citep{yu2018spider, chen2021evaluating, austin2021program, zan2022large}, with HumanEval being the general standard for benchmarking new models on code synthesis. However, current models, aided by inference techniques, can achieve up to a 94\% solve rate on HumanEval \citep{zhou2023language}, indicating a need for harder, more complex, yet self-contained coding tasks to probe the upper limit of code reasoning. Competitive programming questions have thus been proposed as a more difficulty evaluation metric, with most problems coming from online platforms such as Codeforces, Atcoder, Kattis...\citep{li2023taco, huang2023competition, jain2024livecodebench, hendrycks2021measuring, li2022competition}. However, these problems largely do not contain quality problem analyses, comprehensive correctness-defining test cases, with many problems being presented purely symbolically. This means that it can only weakly evaluates the model's ability to creatively reason in grounded problem environments, a crucial ability of well-rounded reasoners.
\paragraph{Inference Time Methods for LMs} Inference time methods have seen significant success in improving reasoning capabilities by conditioning generations on environment feedback, task-specific knowledge, natural language reflections, and planned summaries \citep{shinn2023reflexion, chen2023teaching, madaan2023self, yao2022react, zelikman2022parsel, zhou2023language, gao2023retrieval, le2022coderl}. However, their utility on code domains has only previously been explored on simple program synthesis tasks such as HumanEval and MBPP \citep{chen2021evaluating, austin2021program}. In this paper, we additional provide insights on their performance in competitive programming, a much more difficult domain. Our instantiation of retrieval augmented generation additionally takes inspiration from cognitive architectures for humans reasoning \citep{sumers2023cognitive} and classical case-based reasoning literature \citep{10.5555/196108.196115, 10.5555/538776}, mirroring the types of information humans find useful for problem solving. 
\paragraph{Human Model Interaction} \cite{sumers2022talk} investigates agent learning from human provided feedback under synthetic tasks. \cite{macina2023mathdial} aims to provide a tutoring ruleset to effectively engage LMs in dialogue math problem solving. In this paper we adopt a similar setup to code, applying a specified interaction ruleset to gauge the ability of models to respond to feedback.

\section{The USACO Benchmark}
The USACO benchmark consists of 307 high-quality expert-written problems from past USA Computing Olympiad contests (\url{https://usaco.org}).
Each problem consists of a problem description with instructions for reading and writing from standard input and output;  0-2 sample tests; 10-17 hidden tests verifying solution correctness; time and memory limits verifying solution complexity; and an official human-written \textit{problem analysis} explaining the solution in detail with corresponding Python code.

\paragraph{Problem Difficulty} Problems are divided into tiers of increasing difficulty consisting of bronze, silver, gold, and platinum. At all levels, solutions typically require ad hoc algorithmic reasoning and, unlike interview-level problems, rarely follow directly from well-known algorithms. Gold and platinum problems may additionally require knowledge of known algorithms and data structures, often using them in unorthodox ways.
\paragraph{Task Formulation} A model is given the problem description, including any available samples, and time and memory limits. The model must then produce a code solution, which is run by the judge and accepted if it produces the expected outputs on all hidden tests under the given limits, enforcing both correctness and the desired asymptotic efficiency. Note that the model cannot get access to hidden test inputs or outputs, but can receive information on how many tests a given solution has passed.

\begin{table}[t]
\vspace{-35pt}
\centering
\label{zero-shot-performance}
\begin{tabular}{lccc}
\toprule
\multicolumn{1}{c}{\bfseries Benchmark} & \makecell{{\bfseries Exhaustive} \\ {\bfseries Unit Tests}} & \makecell{{\bfseries Expert-Written} \\ {\bfseries Problem Analyses}} & \makecell{{\bfseries Non-Symbolic} \\ {\bfseries Environments}} \\
\midrule
HumanEval \citep{chen2021evaluating} & \greencheck & \redx & \redx \\
APPS \citep{hendrycks2021measuring} & \redx & \redx & \greencheck  \\
CodeContests \citep{li2022competition} & \redx & \redx & \greencheck \\
USACO (ours) & \greencheck & \greencheck & \greencheck\\
\bottomrule
\end{tabular}
\caption{The USACO benchmark features high-quality problem environments with grounded creative reasoning, complete hidden tests, and expert-written problem analyses with both gold solution reasoning and Python code.}
\vspace{-10pt}
\end{table}
\paragraph{Task Features} We find several features of USACO that make it an effective LLM evaluator. Firstly, USACO problems contain detailed problem environments encouraging grounded, creative reasoning. Problem narratives are high quality, and avoid purely symbolic coding questions such as many presented on platforms like LeetCode or Kattis. Thus, problems focus on a model's ability to reason creatively and ground insights and algorithmic designs to the details of a given scenario.
Furthermore, problems are tiered in difficulty (see Table \ref{benchmark_skills_per_tier}), with bronze tier problems serving as an effective test of pure reasoning, requiring no formal knowledge of data structures and algorithms.
At higher tiers, problems go beyond basic implementation of algorithms, instead requiring creative, grounded algorithm design specific to the desiderata and constraints of the given scenario.
Finally, the USACO benchmark includes expert-written problem analyses containing both natural language and gold Python solutions to all problems, enabling development of rich inference-time techniques and nuanced evaluation beyond unit test execution.
\label{benchmark_skills_per_tier}
\begin{table}[htbp]
\begin{center}
\begin{tabular}{ccccc}
\toprule
\multicolumn{1}{c}{\bfseries Difficulty} & {\bfseries Core Skills Evaluated} \\
\midrule
Bronze & simulation, complete search, sorting, greedy \\
\midrule
Silver & binary search, comparators, graphs, trees, \\
& floodfill, prefix sums, bitwise operators \\
\midrule
Gold & dynamic programming, disjoint set union, \\
& spanning trees, Euler tour, combinatorics \\
\midrule
Platinum & segment tree, range queries, binary jumping, \\
& sweep line, convex hull, flows \\
\bottomrule
\end{tabular}
\end{center}
\caption{Core skills evaluated at each tier of USACO, from \url{https://usaco.guide/}.}
\vspace{-15pt}
\end{table}
\paragraph{Construction: Problems} We collect 484 problems from \url{https://usaco.org} detailing materials on contests hosted between 2011 and 2023 using a custom HTML parser, and use regular expressions to extract wall clock time and memory constraints from problem descriptions. which are manually verified using solutions with correct and incorrect asymptotics.

\paragraph{Construction: Problem analyses}
To assist the development of rich inference-time methods and evaluations, we select the 307 problems out of 484 with full problem analyses. We parse an English-only analysis without code, as well as a ground truth standalone Python 3 code snippet. For the majority of problems where Python code is unavailable, we prompt GPT-4 to translate the code to Python 3 and validate that all code solutions pass hidden tests on the given constraints.

\begin{table}[t]
\label{set_statistics}
\vspace{-35pt}
\begin{minipage}{0.55\linewidth}
\centering
\begin{tabular}{ccccc}
\toprule
\multicolumn{1}{c}{\bfseries Statistics}&  {\bfseries USACO}\\
\midrule
Number of problems & 307 \\
Avg words per problem & 452.9 \\
Avg words per problem analysis & 1107.8 \\
\midrule
Bronze problems & 123 \\
Silver problems & 100 \\
Gold problems & 63 \\
Platinum problems & 21 \\
\bottomrule
\end{tabular}
\end{minipage}
\begin{minipage}{0.45\linewidth}
\centering
\begin{tabular}{lcc}
\toprule
\multicolumn{1}{c}{\bfseries Model} & {\bfseries Pass@1} \\
\midrule
CodeLlama (7B)& 0.19 \\
DeepSeek Coder (7B)& 1.04 \\
GPT-3.5 & 0.59  \\
Claude-3-Sonnet & 2.61 \\
\textbf{GPT-4} & \textbf{8.7}  \\
\midrule
Human Average & 35.83\\
\bottomrule
\end{tabular}
\end{minipage}
\caption{Statistics and zero-shot pass@1 performances on USACO. Human performance estimated through past contest performance.}
\vspace{-10pt}
\label{performance_zero_shot}
\end{table}
\subsection{Baseline Results}
We begin by evaluating zero-shot performance of models representing state-of-the-art coding performance as a baseline: this includes GPT-3.5 (gpt-3.5-turbo-1106), GPT-4 (gpt-4-1106-preview), Claude-3-sonnet, (claude-3-sonnet-20240229), CodeLlama2-Instruct-7B, and Deepseek-Coder-Instruct-7B \citep{roziere2023code, guo2024deepseek, openai2023gpt4}. This is summarized in Table \ref{performance_zero_shot}. Unless otherwise indicated, models were prompted with chain of thought ~\citep{wei2022chain}, refer to figure \ref{fig:zero_shot_prompt} for full prompts. Following previous work on competitive programming \citep{li2022competition, hendrycks2021measuring}, we evaluate primarily based on the unbiased pass@k metric defined in \citep{chen2021evaluating}.

\paragraph{Solve rates near zero for gold difficulty and above.} We find that USACO presents a strong challenge to current generation models. Weaker models like GPT-3.5, CodeLlama, and DeepSeeker cannot solve any problem above silver difficulty, while newer models like GPT-4 have near-zero pass rates for gold difficulty problems and above. For full per-difficulty solve rates, refer to Appendix \ref{difficulty_solve_rates}.
\paragraph{For stronger models, most errors are algorithmic.} Other than CodeLlama, we find that no model errors are significantly due to compilation errors. We detail full results in Appendix \ref{error_breakdown}. This shows at the very least, models are effective in generating syntactically correct code, and indicates more nuanced issues in generations such as problem misunderstandings. We document some samples of problems and their errors in Appendix \ref{fig:USACO_errors} and perform brief qualitative analysis.

\paragraph{Problem release date impacts performance.} We test temporal effects by additionally evaluating models on a small selection of 36 problems released after training cutoff dates. We find that solve rate drops to 0 for all models. However, we do note that USACO questions are well known to increase in difficulty every year, making this likely an effect of difficulty increases, inclusion in pre-training data, as well as small sample size.

\section{Inference Time Techniques for Better Reasoning}
Past work has demonstrated that curated prompting and retrieval strategies can significantly improve performance on various tasks across natural language processing, multi-task QA, and embodied intelligence \citep{yao2022react, wang2023voyager, madaan2023self}. To investigate the effectiveness of such inference-time methods, we adapt self-reflection and retrieval techniques  widely successful in other domains to USACO.

\subsection{Self-Reflection}
Self-reflection techniques aims to allow models to iteratively improve generations by conditioning future output on execution feedback of previous attempts. We primarily experiment with Reflexion \citep{shinn2023reflexion}, a representative technique that additionally maintains an episodic buffer of past attempts to induce better reasoning in future trials.
\paragraph{Setup} The model is first prompted to solve the problem, generating a code solution as well as an explanation of the code. For each iteration of reflection, the model is prompted to first reflect on what went wrong previously, then fix the previous code, given the the execution output of the previous solution, the previous solution itself, as well as the contents of a buffer of past solve attempts. After each iteration, the previous attempt as well as the execution results is added to the buffer. This loop iterates until a maximum number of debugging steps is reached. We set this hyperparameter i = 3 as we observe no empirical gains in solve rate past 3 rounds of debugging: see Appendix \ref{Reflection_hyperparameter_tuning}.
\begin{figure}[t]
\vspace{-35pt}
  \centering
  \includegraphics[width=1\textwidth]{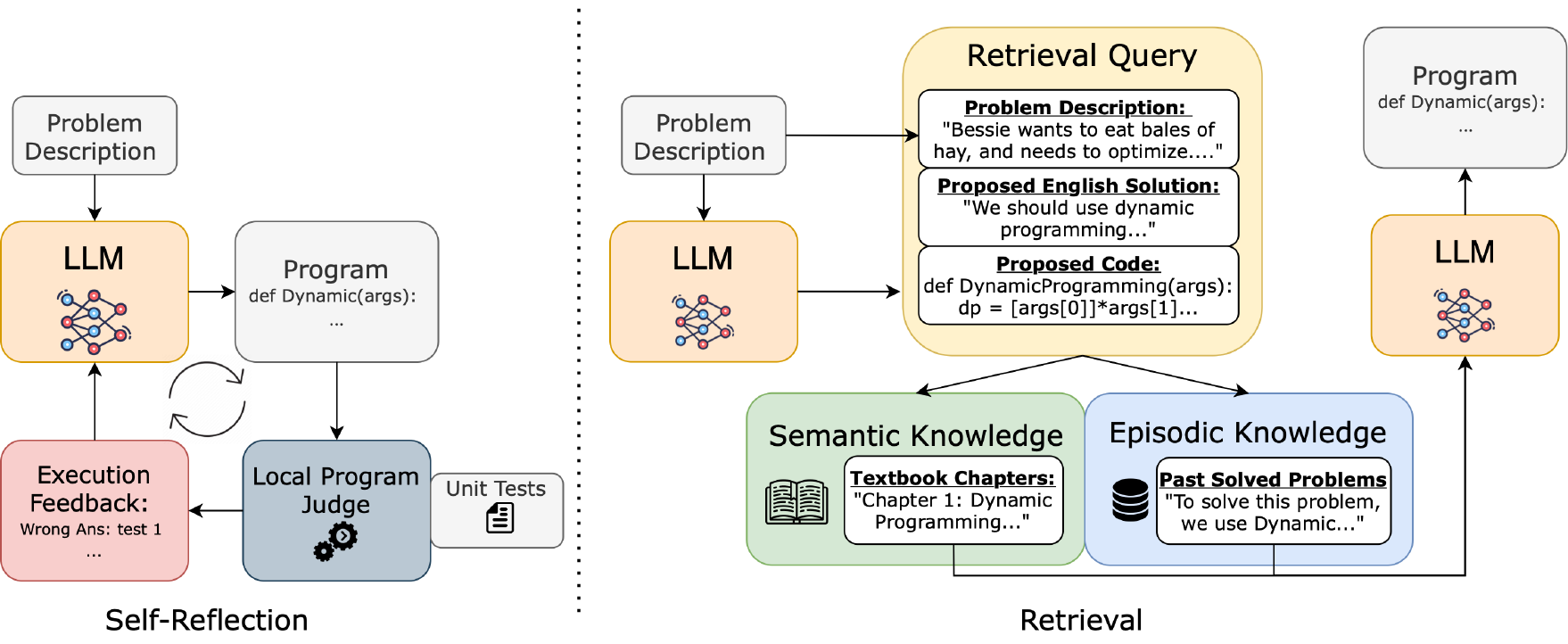} %
  \caption{Overview of inference methods tested: self-reflection (left) prompts the model to review execution feedback to revise its generation; retrieval (right) uses the problem and a draft solution to query relevant semantic and episodic knowledge to generate a more informed final solution.}
  \label{fig:memory_retrieval}
  \vspace{-10pt}
\end{figure}
\subsection{Retrieval Augmented Generation}
Retrieval augmented generation (RAG) has similarly proven to reduce hallucinations and improve reasoning capabilities in a variety of domains. \citep{gao2023retrieval}. However, it has seen limited usage in coding domains as it is difficult to pinpoint what types of knowledge is useful in aiding code generation. We note that when solving problems, humans tend to recall either task-specific, established algorithms, facts, and concepts about the domain, or past experiences and generalizations from solving previous, similar problems relative to the problem at hand. This represents semantic knowledge, and episodic knowledge respectively \citep{sumers2023cognitive}. Inspired by this, we curate two setups we aptly name semantic retrieval and episodic retrieval, and instantiate the semantic knowledge store as a competitive programming textbook, and the episodic knowledge store as the bank of USACO problems and solutions not currently being solved.

\paragraph{Semantic Knowledge Store} We use the cp-algs textbook (\url{https://cp-algorithms.com}), which contains 30 human-written chapters on algorithmic concepts specifically targeted to the USA Computing Olympiad. Chapters contain English text as well as code snippets. Entire chapters barely fit within the context limit of GPT-4: therefore when lower context models (like GPT-3.5), or multiple types of knowledge are incorporated, we truncate the retrieved chapter to fit within the context length.
\paragraph{Episodic Knowledge Store} We simulate a setup where the model has seen all other problems in the USACO set except for the current one it is solving. Thus for each "seen" problem, its corresponding problem description, english solution, and python solution code is concatenated together to form documents to retrieve over. We tune over the number of problems to retrieve, $p$, and find that $p=2$ is optimal for GPT-4, and $p=1$ is optimal for GPT-3.5 and thus report these numbers. Full details can be found in Appendix \ref{hyperparameter_tuning}.

\paragraph{Retrieval Query} Ablations on retrieval queries indicate that the most effective retrieval query utilizes the current problem description, as well as an initial solution attempt containing code and english explanation. This allows accurate retrieve relevant algorithm descriptions from the underlying retrieval corpus, as solely utilizing the problem descriptions does not allow retrieval over algorithmic keywords. We do not count this initial generation as an attempt as it does not get evaluated by our local judge. Additional details on ablation experiments can be found in Appendix \ref{retrieval_query_ablations}. 

\paragraph{Setup} Models are prompted with the problem description, and first generate an initial solution to be used in the retrieval query. This initial solution, along with the problem description, is fed into a BM25 retrieval function, and the highest ranking document is inserted into the context to aid the real solving process. Prompts can be found in Appendix \ref{memory_retrieval_prompts}. We additionally note that retrieval-based methods and reflection methods are orthogonal and can be sequentially applied. Thus, we test settings combining various combinations of retrieval types, as well as retrieval combined with reflection.

\begin{figure}
\vspace{-35pt}
    \centering
    \begin{tabular}{lcccc} %
    \toprule
    \textbf{Technique} & \textbf{GPT-3.5} & \textbf{GPT-4} \\
    \midrule
    Zero-shot & 0.59 & 8.70 \\
     \midrule
    Reflexion & 0.97 & 12.38\\
    Semantic Retrieval & 2.04 & 10.26\\
    Episodic Retrieval & 5.49 & 14.33 \\
     \midrule
    Semantic + Episodic & 6.76 & 12.54\\
    Semantic + Reflexion & 2.64 & 11.82\\
    \textbf{Episodic + Reflexion} & \textbf{8.79} & \textbf{20.20} \\
     \midrule
    Semantic + Episodic + Reflexion & 7.52 & 18.05 \\
    \bottomrule
    \end{tabular}
    \label{inference_performances}
    \caption{Pass@1 Performance of Reflexion and Retrieval based methods on USACO.}
    \vspace{-10pt}
\end{figure}
\paragraph{Evaluation} All methods are evaluated Pass@1.For self-reflection, we adapt methodology in \cite{shinn2023reflexion}, and provide only the execution results of the exposed sample test cases for models to reflect over. GPT-3.5 and GPT-4 were used for initial experiments: future work will involve expanding results on open source models.

\subsection{Results}
We summarize model performances under each setting in table \ref{inference_performances}. Combining episodic retrieval and reflexion maximizes performance gains by well over 
\textbf{doubling} the performance of zero-shot GPT-4. We condense key findings below:
\paragraph{Episodic Retrieval works across model sizes, unlike Reflexion.} We find that the ability to self-reflect effectively is an emergent property of stronger models, consistent with \citep{shinn2023reflexion, chen2023teaching}. However, both semantic and episodic retrieval are still effective, with episodic retrieval even causing \textbf{GPT-3.5 to approach GPT-4's zero-shot performance}. This is likely because self-reflection relies on the internal model's strength to reason over sparse, binary reward signals. Retrieval, on the other hand, allows models to reference existing reasoning and code snippets, requiring less intrinsic model capabilities. Our findings thus corroborate \cite{li2023explaining}, where LMs can understand much more complex competitive programming solutions than they can produce.
\paragraph{Episodic Retrieval and Reflexion have strong synergy.} Episodic Retrieval reaches new maximums when combined with Reflexion, but not with Semantic Retrieval. We find that for GPT-4, \textbf{70.2\%} of newly solved problems (relative to zero-shot) with semantic retrieval are also newly solved by episodic retrieval. It provides one possible explanation as to why combining the two may decrease performance: the additional knowledge provided by our implementation of semantic retrieval trades off against its long contexts, which current LLMs are known to be struggle with \citep{liu2024lost}. In contrast, only \textbf{45.9\%} of newly solved problems with Reflexion overlap with episodic retrieval, pointing to better synergy.

\paragraph{Platinum problems remain unsolved.} Although solve rates on gold problems grow significantly, platinum problems remain unsolved, posing an open challenge for future inference techniques and foundation models. For more details, refer to Appendix \ref{per_difficulty_solve_rates}
\paragraph{Performance gains are not due to memorization.} A competing hypothesis for the success of retrieval is that adding retrieved solutions increases memorization effects for the problem currently being evaluated, rather than the model critically engaging with the content of the retrieved content itself. To test for this, we remove critical sections of retrieved solutions and find significant drops in performance. In addition, qualitative analysis indicates no significant overlap in generated and official published solutions. Full experiment details can be found in Appendix \ref{addressing_memorization}.
\begin{figure}[t]
\vspace{-35pt}
  \centering
  \includegraphics[width=1\textwidth]{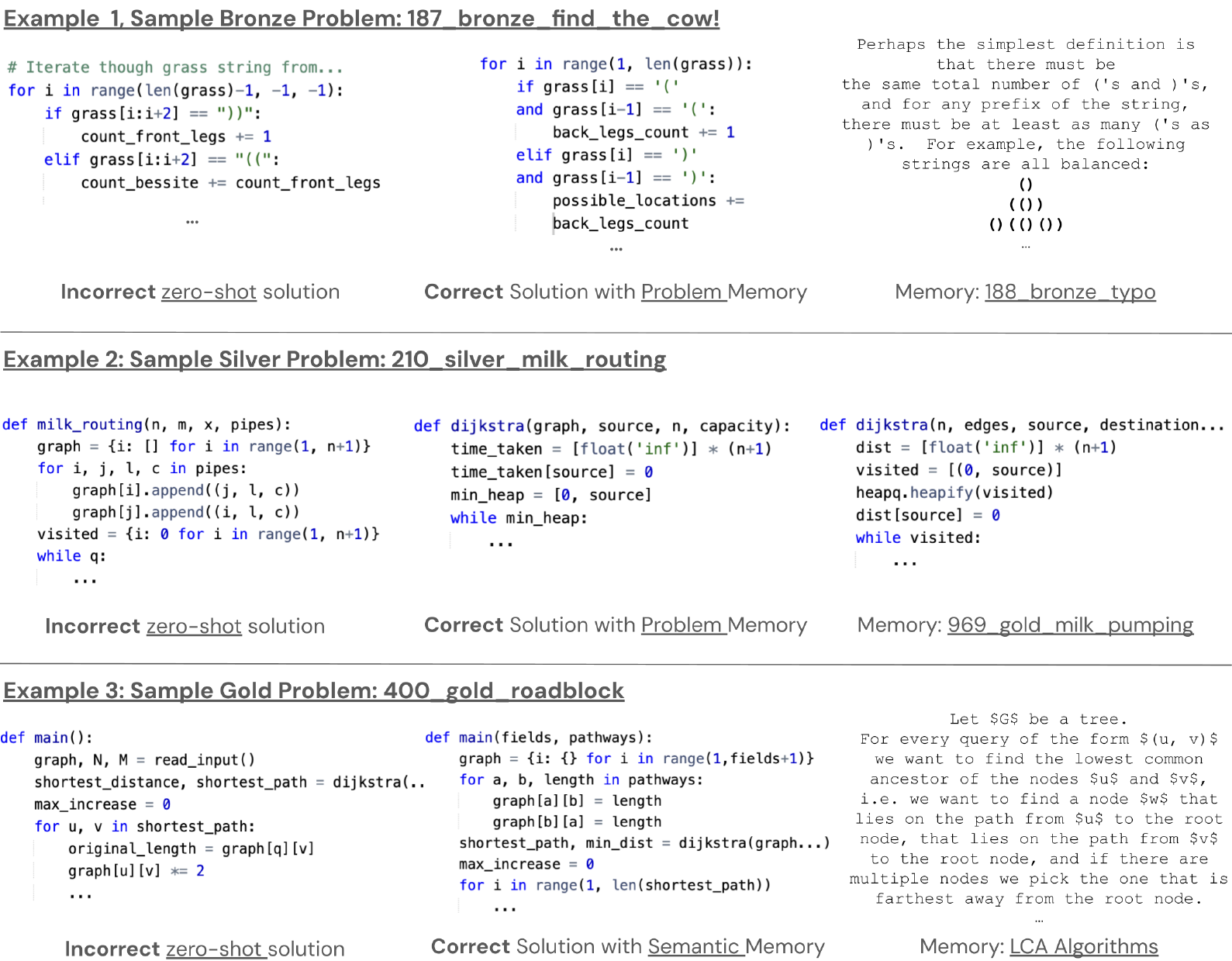} %
  \caption{Three examples of problems previously unsolved, but solved with retrieval.}
  \label{fig:qual_analysis_inf_method}
  \vspace{-10pt}
\end{figure}
\subsection{Qualitative Analysis}
We focus our qualitative analysis on the improvements of retrieval, our individually strongest inference-time technique. Here, we isolate 3 examples of problems newly solvable with RAG in figure \ref{fig:qual_analysis_inf_method}.
\paragraph{Example 1: Models can borrow reasoning about similar problem environments} Here, both the central problem and the most relevant retrieved problem requires models to parse and operate over parenthesis-only strings. The retrieved solution + code thus provides models with sample reasoning over this tricky and mistake-prone problem environment, allowing models to generate more accurate code.
\paragraph{Example 2: Models can adopt existing code structure and algorithms} In this example, both problems require the use of Dijkstra's shortest path algorithm. However, the initial implementation contains many small, nuanced bugs due to its attempts to push the entire solution into a single function. This not only makes the code less modular, but it also leaves greater room for error. With episodic retrieval, we see the model grounding itself in the retrieved content, borrowing significant code structure and algorithmic reasoning of the retrieved problem.
\paragraph{Example 3: Models can utilize algorithmic concepts and reasoning from texts} The given problem here requires finding a shortest path within a grid containing roadblocks. We see that the model fetches a textbook chapter on lowest common ancestor algorithms, which covers tangentially applicable graph traversal techniques to the problem. Interestingly enough, it does not fetch the chapter on Dijkstra's algorithm for shortest paths, the algorithm utilized in the official problem analysis. Visual inspection of the Dijkstra's chapter indicated that the chapter was short and light on details, thus receiving a low retrieval score. This highlights the flexibility of the retrieval query to adapt to low quality documents, finding suitable replacements.

\section{Human-in-the-loop Guidance}
In benchmark evaluations, we found a wide diversity in the distribution of model errors: from problem misunderstandings, to subtle off-by-one implementation issues. To further examine how far a model is from solving a given problem, we perform a human study via an interactive ``tutoring setup."
\paragraph{Setup} A human provided with problem solutions engages in a multi-turn exchange with a model, at each step providing feedback on mistakes under a specified ruleset. Notably, the human is not allowed to provide specific fixes rather only general instructions (e.g., ``you are using the wrong algorithm" or ``this code may result in index out of bounds"). The full interaction ruleset is detailed in Appendix \ref{human_tutoring_prompt}. The model is then prompted to fix its mistakes: we allow a maximum of 1 code execution to simulate a pass@1 setup, a maximum of 5 code generations to limit conversation length, and a total of 3 attempts per problem.

\begin{wrapfigure}{r}{0.35\textwidth} 
  \centering
    \vspace{-10pt}
  \includegraphics[width=0.35\textwidth]{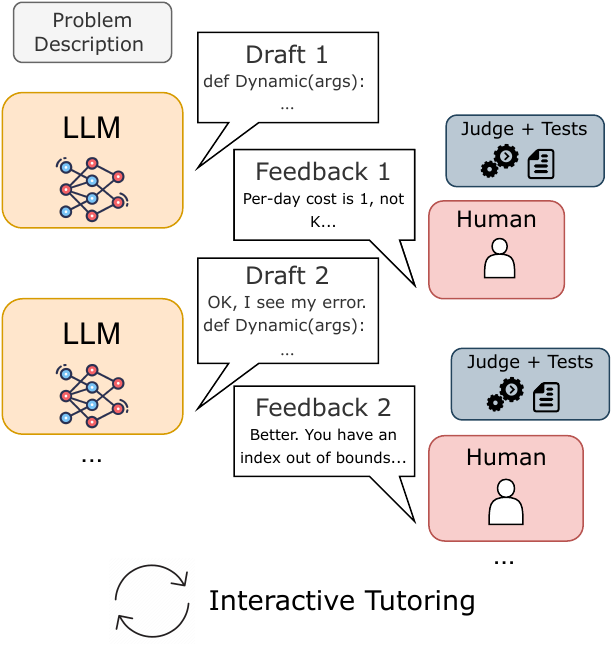}
  \vspace{5pt}
  \begin{tabular}{lc}
    \toprule
    Model & \% Solved \\ 
    \midrule
    GPT-3.5 & 0\% \\
    GPT-3.5 + tutoring & 0\% \\
    GPT-4 & 0\% \\
    \textbf{GPT-4 + tutoring} & \textbf{86.7\%} \\
    \bottomrule
  \end{tabular}
  
  \caption{Human-in-the-loop interactive ``tutoring" setup: GPT-4 successfully incorporates feedback while GPT-3.5 does not.}
  \label{fig:human_in_loop}
  \vspace{-20pt}
\end{wrapfigure}
\paragraph{Results} Surprisingly, we find that on a small set of 15 problems\footnote{13 bronze, 1 silver, 1 gold} on which GPT-3.5 and GPT-4 achieve zero pass rate using all of the above inference-time methods, the human-in-the-loop setup raises GPT-4 performance from 0\% to 86.7\% (13 problems solved) while not improving GPT-3.5 performance from 0\%. Qualitatively, GPT-3.5 consistently hallucinates fixes irrelevant to feedback, while GPT-4 pinpoints accurate algorithmic fixes in response to minimalistic feedback. A sample trajectory is provided in Appendix \ref{human_sample_trajectories}.
\paragraph{Problem-level Analysis} We found that GPT-4 was more responsive to blanket feedback that its algorithm or understanding of an environment concept was incorrect, and more capable of landing on the correct strategy in its second attempt, whereas GPT-3.5's retries are typically similarly unhelpful. For example, GPT-4 lands on the correct overall solution strategy after being instructed to not use a heap in ``Hungry Cow," and similarly after being instructed not to use DP in ``The Lost Cow." In ``Photoshoot," it suffices to ask GPT-4 ``Is there any way we can use the inherent ordering of the cows and directly calculate the number of steps necessary?" to correct it from a simulation to a precomputed parity-counting approach; explaining this to GPT-3.5 does not yield any progress. Similarly, GPT-4 is also able to follow instructions on specific fixes, unlike GPT-3.5. For example, on ``The Lost Cow," GPT-4 tends to calculate the next position using the current position instead of the \textit{initial} position -- if given this fact, it produces the correct implementation.

\paragraph{Discussion} Our human-in-the-loop results highlight that the full capabilities of models may not be captured by solve rate; among two models failing on a given problem, one may be one correction away from a fully correct solution, whereas the other may fundamentally misunderstand the problem scenario. This motivates better evaluation metrics beyond execution success (pass@k). Another perspective on our results is that GPT-4 has further reasoning capabilities that may be ``unlocked" by human-level corrective feedback, highlighting the need for better methods to generate such feedback.

\section{Conclusion}
 In this paper, we introduce the USACO benchmark for rigorously evaluating code language models on tasks involving grounded ad hoc reasoning and novel algorithmic thinking. We observe that foundation models previously shown to excel at basic coding tasks like HumanEval \citep{chen2021evaluating} perform poorly zero-shot in these more challenging scenarios, but that providing models with task-specific knowledge stores can well over double zero-shot performance. We hope that our evaluation of current models' limitations and findings on the effectiveness of semantic and episodic knowledge help lay groundwork for its integration into future models and language agents alike.

\section{Reproducibility}
We release all data and code at \url{https://princeton-nlp.github.io/USACOBench/}. We advise others to use isolated execution environments when reproducing experiments as the generated code is not validated before execution.
\bibliography{colm2024_conference}
\bibliographystyle{colm2024_conference}

\appendix
\section{USACO Details}
\label{USACO}
\subsection{USACO Zero-Shot Prompt}
\begin{figure}[!htbp]
  \centering
  \includegraphics[width=1\textwidth]{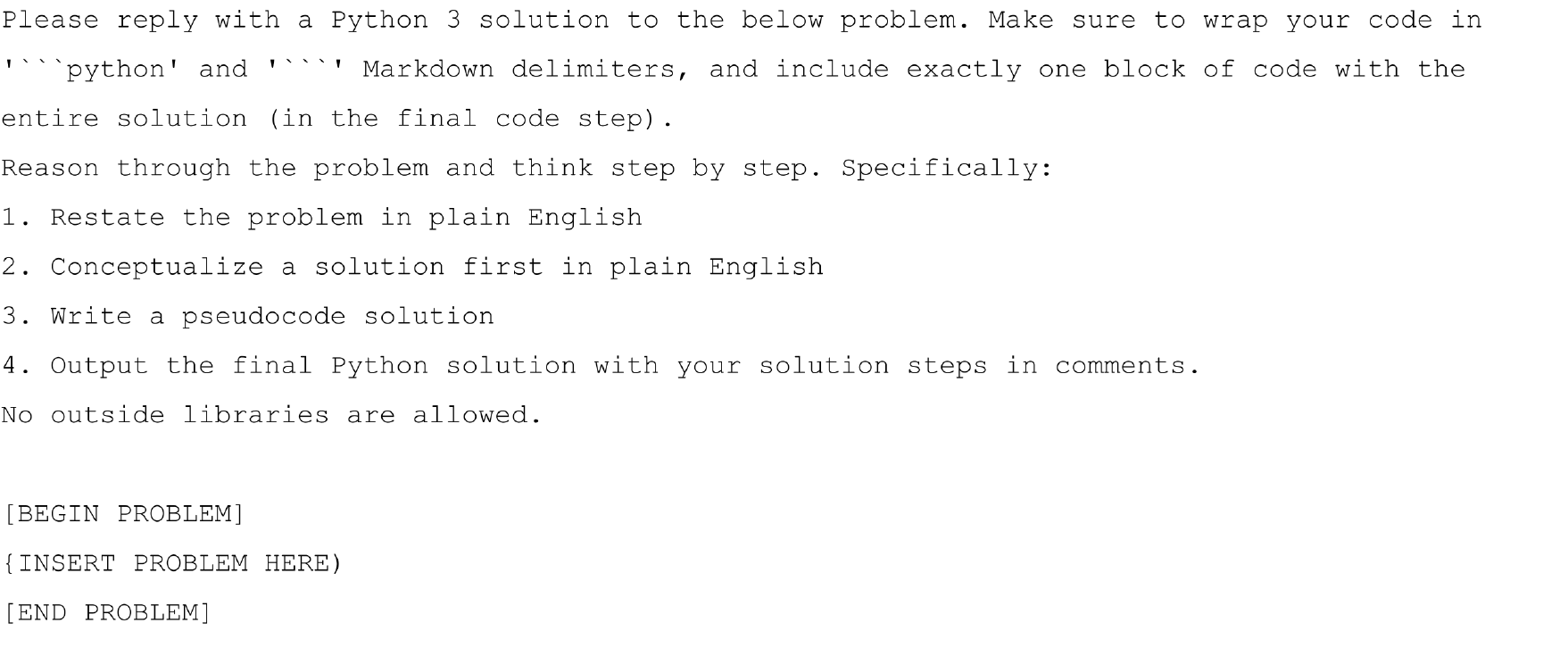} %
  \caption{USACO zero-shot prompt. We find that asking the model to describe its reasoning + pseudocode first lead to more parse-able and human-interpretable generations, making qualitative analysis greatly simplified. It is also beneficial for utilization in retrieval queries, as it contains more algorithmic keywords. However, it does not boost the performance significantly: the choice to prompt it to do the extra steps for zero-shot performance evaluation is more for interpretability benefits.}
  \label{fig:zero_shot_prompt}
\end{figure}
\pagebreak
\subsection{USACO Zero-Shot Error Breakdown}
\begin{table}[!htbp]
\label{error_breakdown}
\centering
\begin{tabular}{lccccccc} %
\toprule
\textbf{Model} & \textbf{Wrong Ans.} & \textbf{TLE} & \textbf{MLE} & \textbf{Runtime} & \textbf{Syntax + Other} \\
\midrule
CodeLlama (7B)& 61.10 & 5.05 & 0 & 9.89 & 23.85 \\
DeepSeek Coder (7B)& 80.62 & 10.04 & 0 & 6.82 & 1.86 \\
Claude-3-Sonnet & 55.70 & 24.42 & 0 & 14.66 & 2.61\\
GPT-3.5 & 77.42 & 7.02 & 0 & 13.37 & 1.82 \\
GPT-4 & 44.15 & 38.20 & 0 & 10.41 & 1.18 \\
\bottomrule
\end{tabular}
\caption{Error distributions zero-shot, in \%. TLE indicates time limit exceeded, and MLE indicates memory limit exceeded. "Other" generally represents errors stemming from models outputting incorrectly formatted code.}
\end{table}
\subsection{USACO Zero-Shot Performance By Difficulty}
\label{difficulty_solve_rates}
\begin{table}
\centering
\begin{tabular}{ccccc}
\toprule
 \multicolumn{1}{c}{\bfseries Model} & {\bfseries Bronze} & {\bfseries Silver} & {\bfseries Gold} & {\bfseries Platinum} \\
\midrule
CodeLlama (7B)& 0.41 & 0 & 0 & 0  \\
DeepSeek Coder (7B)& 2.30 & 0 & 0 & 0  \\
GPT-3.5 & 1.46 & 0 & 0 & 0  \\
Claude-3-Sonnet & 5.69 & 1.00 & 0 & 0 \\
GPT-4 &\textbf{19.11} & \textbf{3.10} & \textbf{0.16} & 0  \\
\bottomrule
\end{tabular}
\caption{Zero-shot performance of models by difficulty}
\end{table}
\pagebreak
\subsection{USACO Qualitative Analysis}
\begin{figure}[!htbp]
  \centering
  \includegraphics[width=1\textwidth]{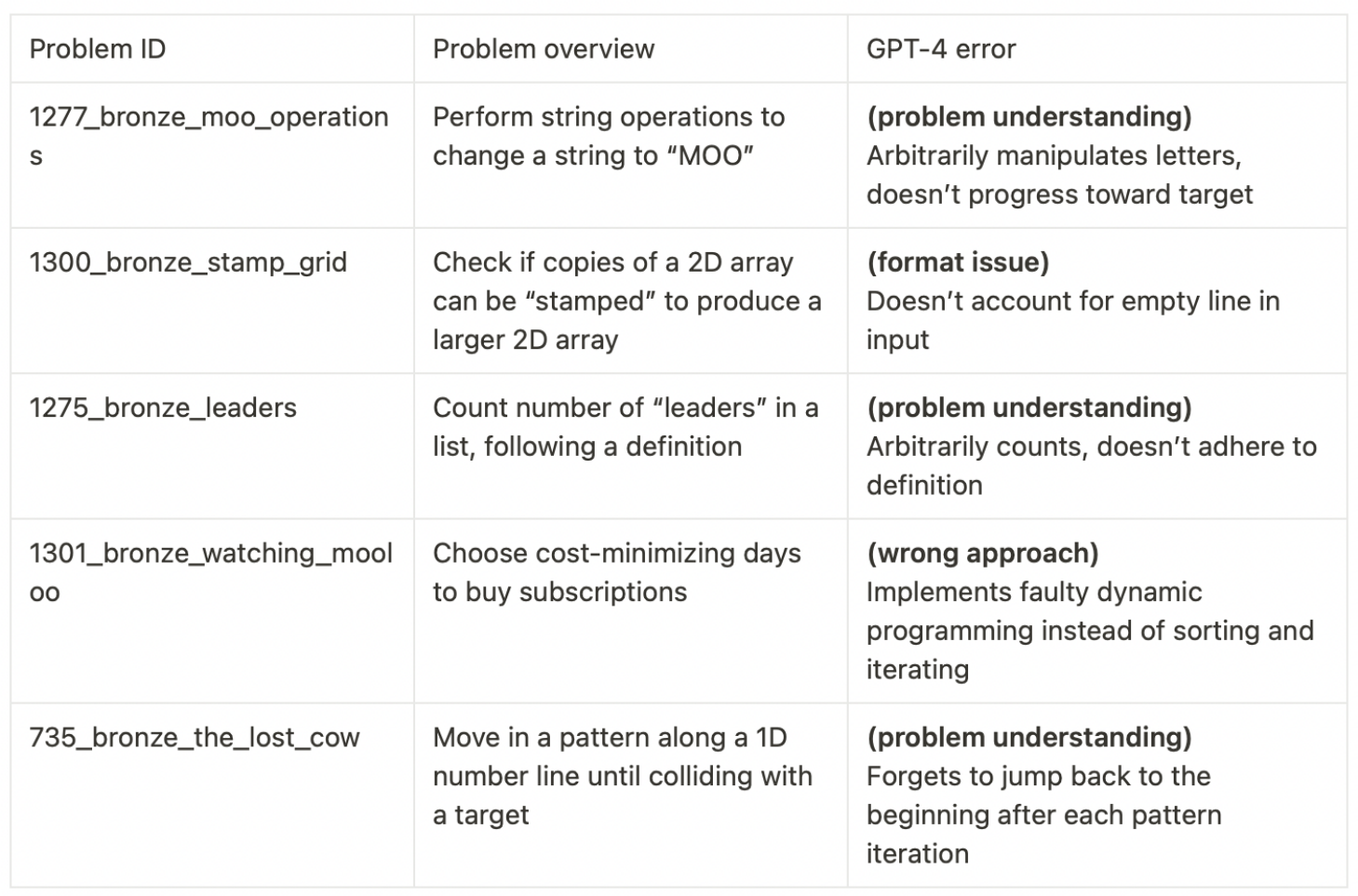} %
  \caption{An overview of some common, general errors on bronze problems, as easier problems generally have only 1-2 points of failure and is simpler to analyze.}
  \label{fig:USACO_errors}
\end{figure}
\subsection{Sample Generation}
\begin{figure}[!htbp]
  \centering
  \includegraphics[width=1\textwidth]{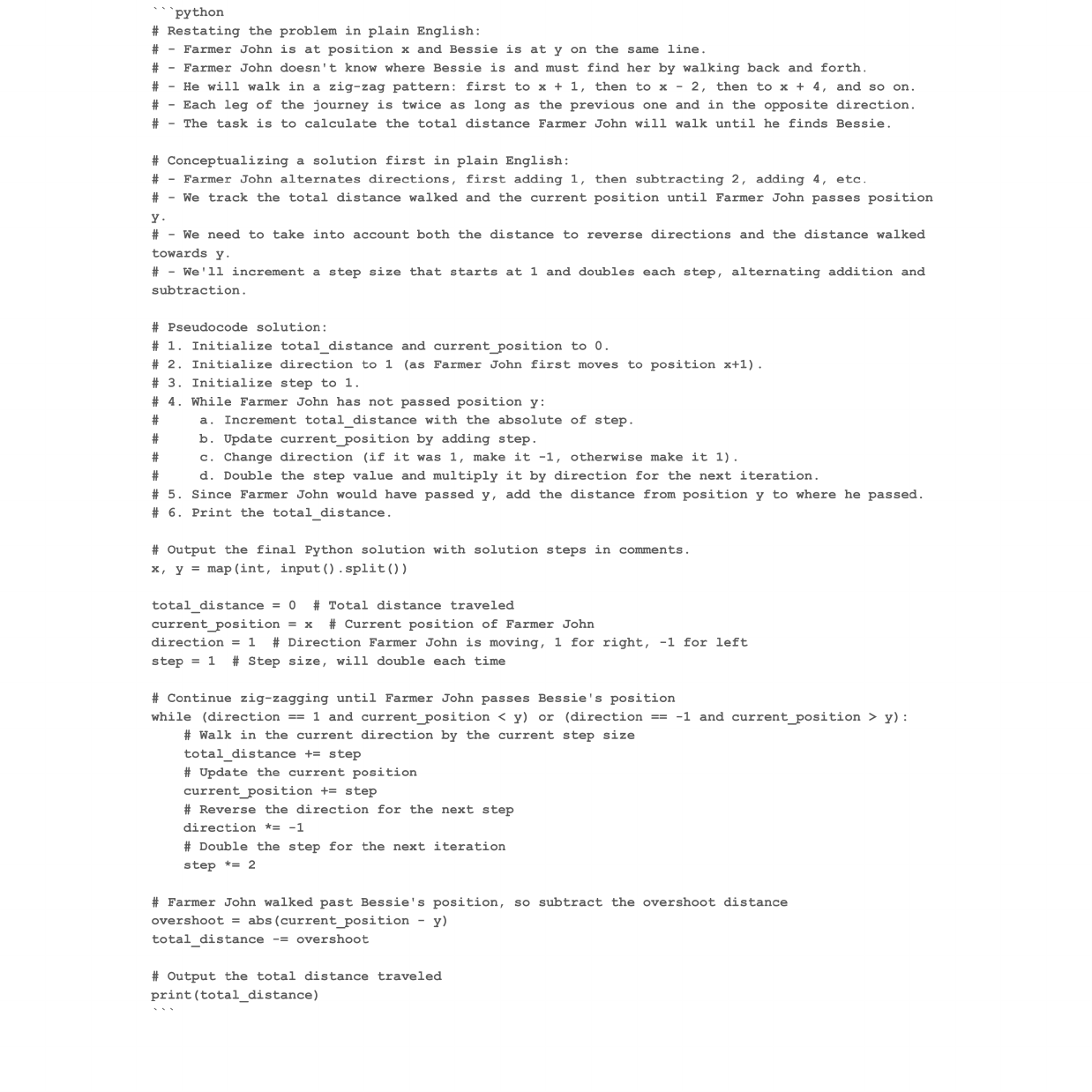} %
  \caption{Sample zero-shot generation by GPT-4}
  \label{fig:sample_generation}
\end{figure}
\pagebreak
\section{Inference Time Methods}
\subsection{Addressing Memorization}
We address memorization by performing ablations on the corpus in table \ref{memorization}: if retrieved problem solutions were spurring regurgitation of memorized solutions to the current problem, removing core parts of the retrieved solutions should not dwindle this effect. However, we find that it does: utilizing solely the problem description only retains 3.18\% of the performance, suggesting that models are truly utilizing the reasoning of similar problems provided in context to inform their generations.
\label{addressing_memorization}
\begin{table}[!htbp]
\begin{center}
\begin{tabular}{ccccc}
\toprule
\multicolumn{1}{c}{\bfseries Retrieval Content} & {\bfseries \% of maximum performance}\\
\midrule
\textbf{PD} & \textbf{3.18} \\
PD + Code + English Solution & 100 \\
\bottomrule
\end{tabular}
\end{center}
\label{memorization}
\caption{Ablations on episodic corpus content. PD represents problem description.}
\end{table}

\subsection{Per Difficulty Pass Rates}
\label{per_difficulty_solve_rates}
\begin{table}[!htbp]
\label{difficulty_breakdown}
\centering
\begin{tabular}{lccccccc} %
\toprule
\textbf{Model} & \textbf{Bronze} & \textbf{Silver} & \textbf{Gold} & \textbf{Platinum} \\
\midrule
Reflexion & 24.39 & 5 & 4.76 & 0 \\
Semantic Retrieval & 19.72 & 6.0 & 1.98 & 0\\
Episodic Retrieval & 26.22 & 9.5 & 3.6 & 0 \\
Episodic Retrieval + Reflexion & 35.77 & 14.0 & 6.35 & 0\\
\bottomrule
\end{tabular}
\caption{GPT-4 Pass@1 rates for inference methods per difficulty: We report incrementally superior methods.}
\end{table}

\subsection{Retrieval Query Ablations}
\label{retrieval_query_ablations}
\begin{table}[htbp]
\begin{center}
\begin{tabular}{ccccc}
\toprule
\multicolumn{1}{c}{\bfseries Query Content} & {\bfseries Performance}\\
\midrule
PD & 13.36 \\
PD + Attempted Code Solution & 14.33 \\
PD + Attempted English + Code Solution & 14.98 \\
\bottomrule
\end{tabular}
\end{center}
\caption{Ablations on retrieval query, PD represents problem description. Performance measured in pass@1. We find that generally most retrieval queries are somewhat effective, however, including code attempts as well as english solution performs the best, as it allows the maximum matching of relevant keywords between compared documents.}
\end{table}
\pagebreak
\subsection{Environment duplicates}
\begin{figure}[!htbp]
  \centering
  \includegraphics[width=1\textwidth]{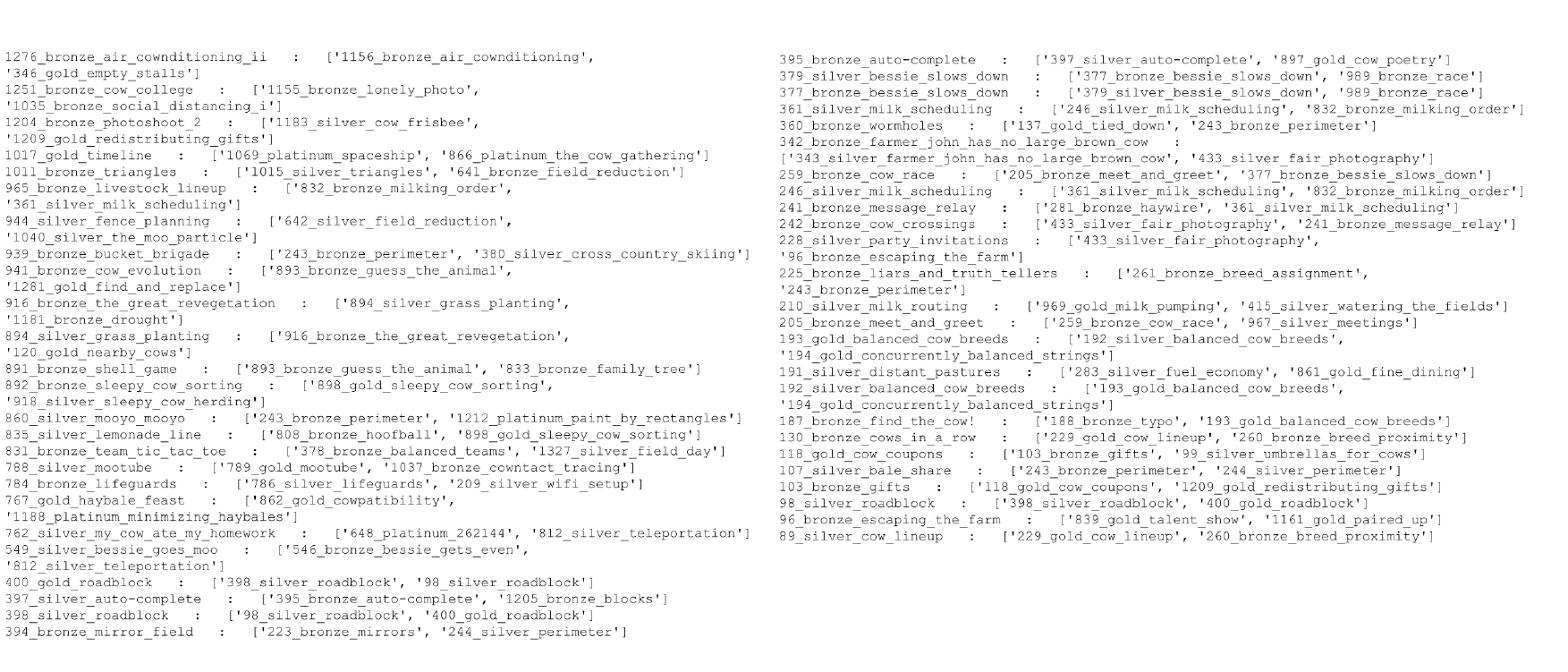} %
  \caption{A list of problems that were not originally solved zero-shot, but is solved with episodic retrieval, as well as a list of the relevant problems retrieved by problem id, listed in decreasing order of relevancy.}
  \label{fig:retrieved_problems}
\end{figure}
\paragraph{Problem environment duplicate retrieval} We want to quickly point out here that USACO reuses problem environments occasionally. For example, there exists both gold and platinum versions of the problem "pareidolia," both asking users to develop different algorithms regarding to strings. Although the problems utilize the same problem environment, the algorithms and reasoning behind the solutions differ greatly, making it still a nontrivial task to solve the problem even given the solutions to the alternate problem. Additionally, we find that only 33\% of newly solved problems contain one or more retrieved content that fall into this category: the full list can be found in figure \ref{fig:retrieved_problems}. The rest retrieve problems that are completely separate.
\subsection{Hyperparameter Tuning}
\label{hyperparameter_tuning}
\begin{table}[!htbp]
\label{ER_hyperparameter_tuning}
\centering
\begin{tabular}{cc}
\toprule
\multicolumn{1}{c}{\bfseries Problems Retrieved} & {\bfseries Pass@1} \\
\midrule
$p=1$ & 13.03 \\
$p=2$ & 14.33 \\
$p=3$ & 13.11 \\
$p=4$ & 12.38 \\
\bottomrule
\end{tabular}
\caption{Episodic retrieval hyperparameter tuning: Here we tune over how many problems to retrieve over on the USACO307 dataset. GPT-4-turbo-1106 was used in all experiments here. We see that $p=2$ is optimal in pass@1. We did not test resampling for greater numbers of $p$ to conserve budget as performance in pass@1 was already dropping.}
\end{table}

\begin{table}[!htbp]
\label{Reflection_hyperparameter_tuning}
\centering
\begin{tabular}{cc}
\toprule
\multicolumn{1}{c}{\bfseries Reflexion Iterations} & {\bfseries Pass@i} \\
\midrule
$i=0$ & 8.7 \\
$i=1$ & 10.75 \\
$i=2$ & 12.28 \\
$i=3$ & 12.38 \\
$i=4$ & 12.38 \\
$i=5$ & 12.40 \\
\bottomrule
\end{tabular}
\caption{Reflection iteration tuning: Here we tune over how many times to iterate. All experiments were done with GPT-4. $i=0$ indicates the original solve rate without any reflection. We see that solve rates remain relatively static after 3 iterations.}
\end{table}
\pagebreak
\subsection{Reflexion Prompt}
\begin{figure}[htbp]
  \centering
  \includegraphics[width=1\textwidth]{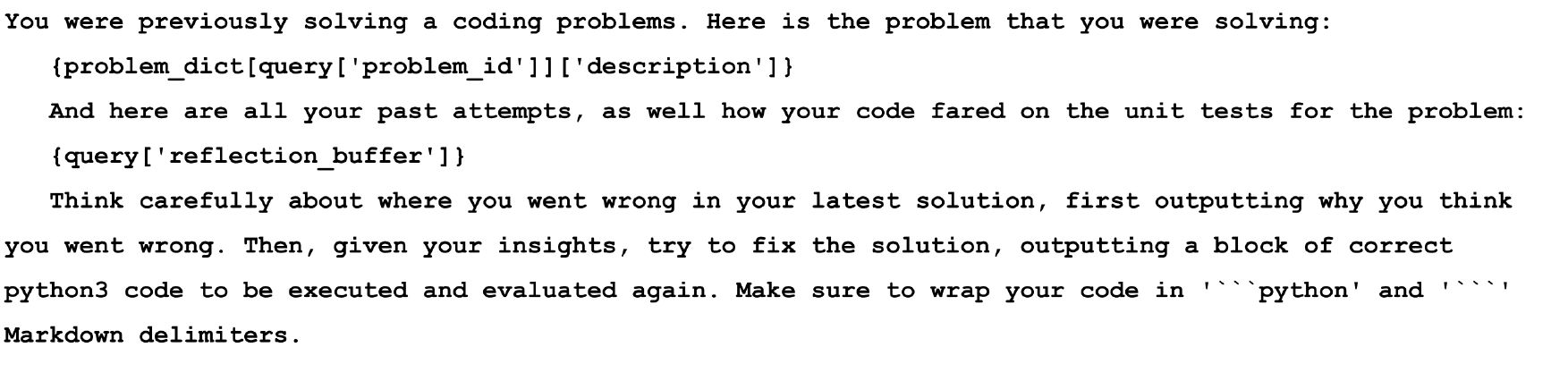} %
  \caption{Reflexion prompt}
  \label{fig:reflexion_prompt}
\end{figure}
\subsection{Retrieval Prompts}
\label{memory_retrieval_prompts}
\begin{figure}[htbp]
  \centering
  \includegraphics[width=1\textwidth]{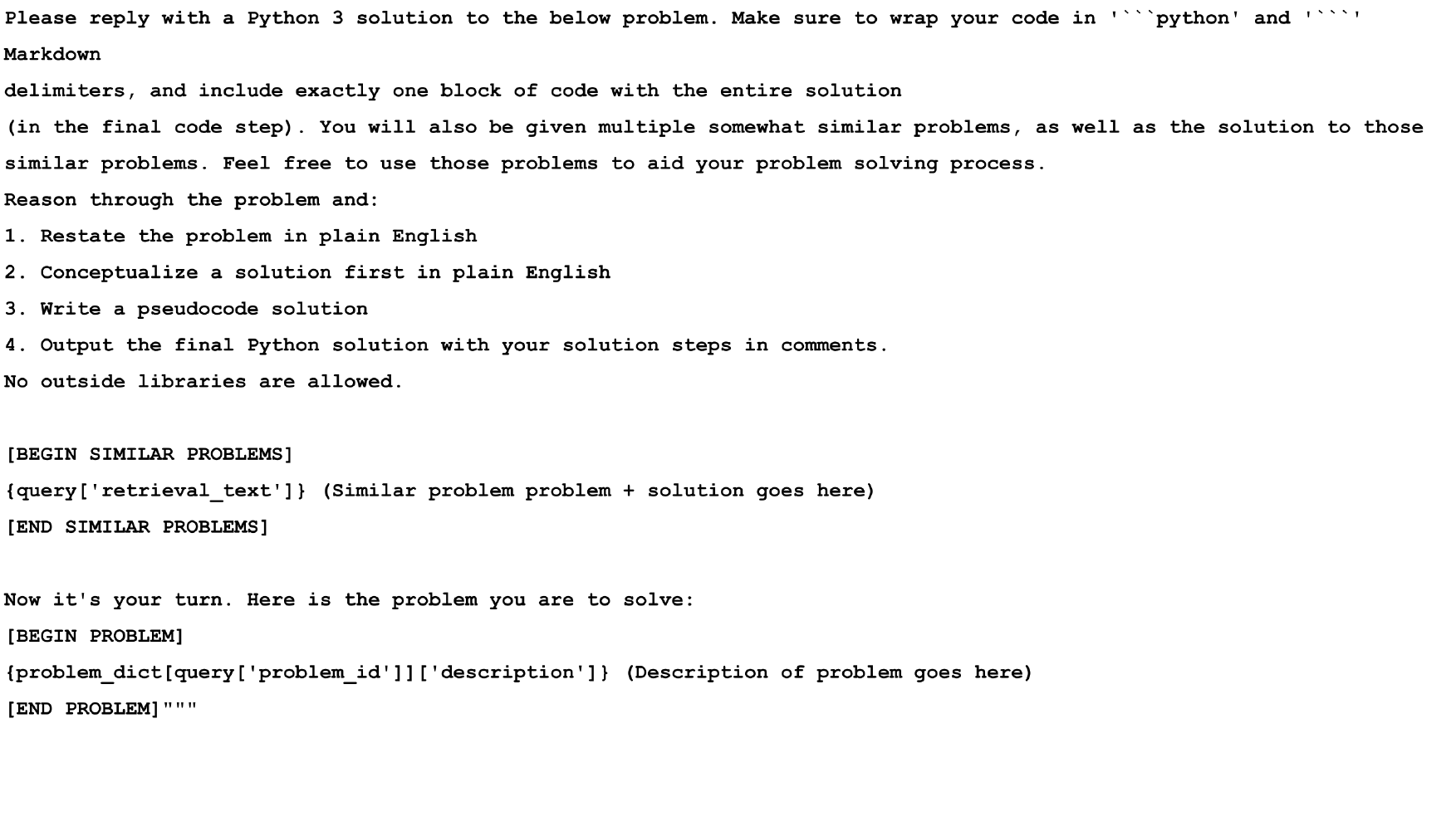} %
  \caption{Episodic Retrieval Prompt}
  \label{fig:episodic_retrieval_prompt}
\end{figure}
\begin{figure}[htbp]
  \centering
  \includegraphics[width=1\textwidth]{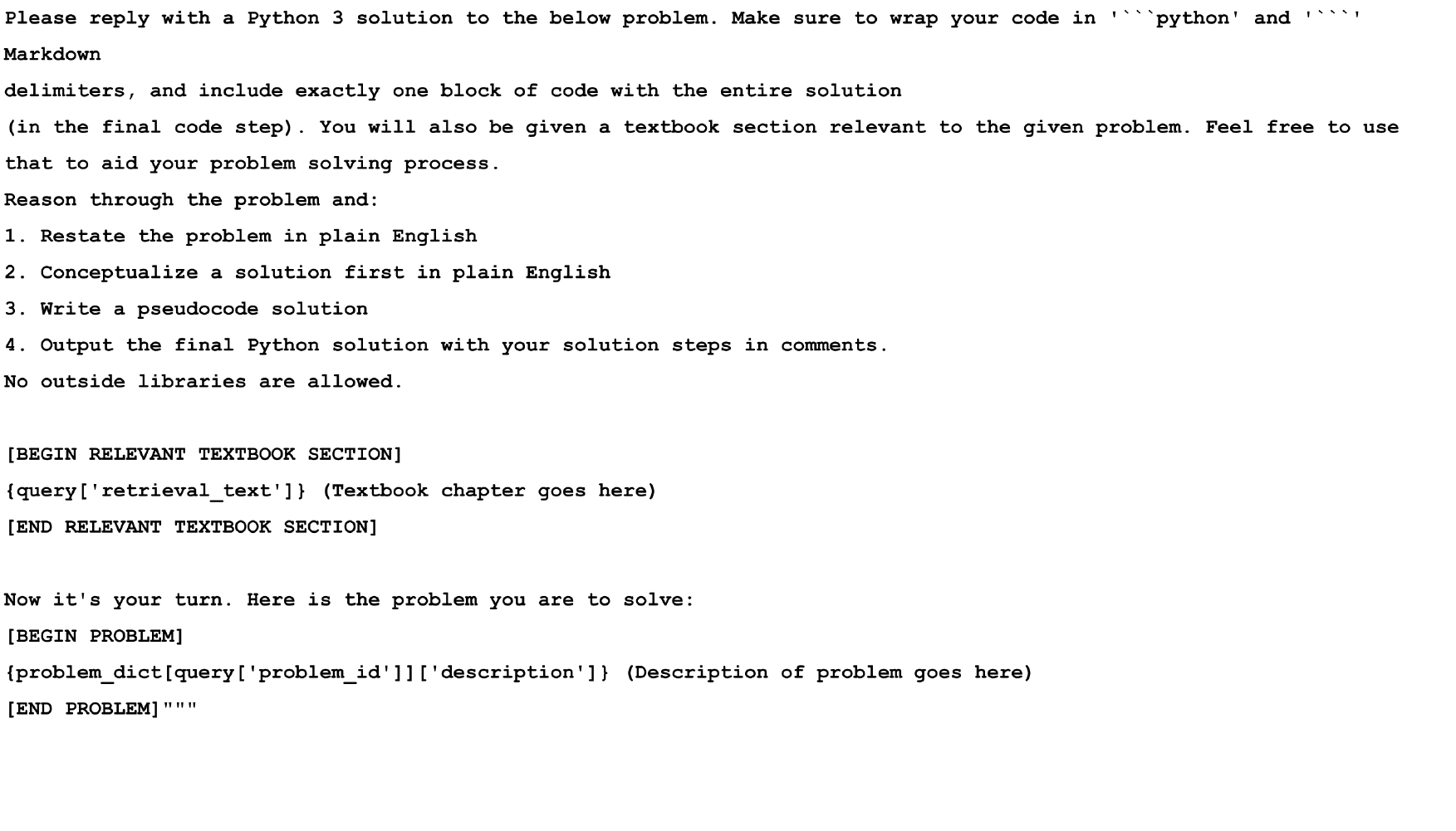} %
  \caption{Semantic Retrieval Prompt}
  \label{fig:semantic_retrieval_prompt}
\end{figure}

\begin{figure}[htbp]
  \centering
  \includegraphics[width=1\textwidth]{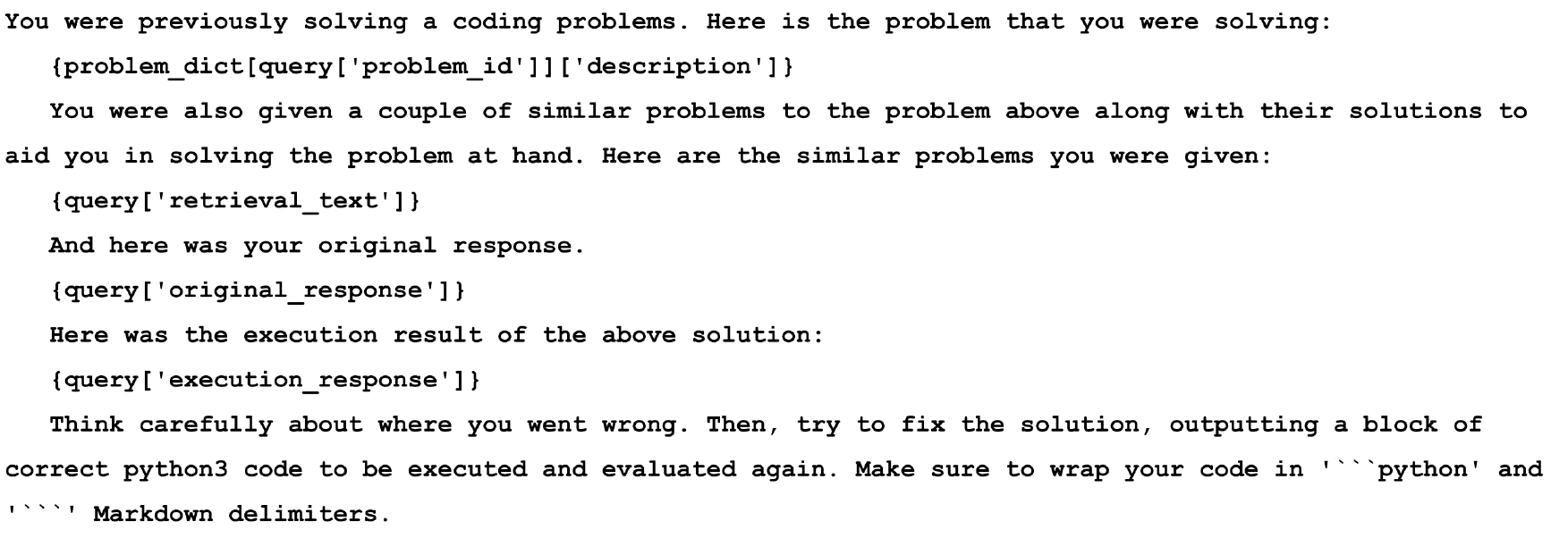} %
  \caption{Reflexion + Retrieval Prompt}
  \label{fig:reflexion_memory_retrieval_prompt}
\end{figure}

\pagebreak
\section{Human-in-the-loop}
\subsection{Human-in-the-loop Prompt}
\label{human_tutoring_prompt}
\begin{figure}[htbp]
  \centering
  \includegraphics[width=1\textwidth]{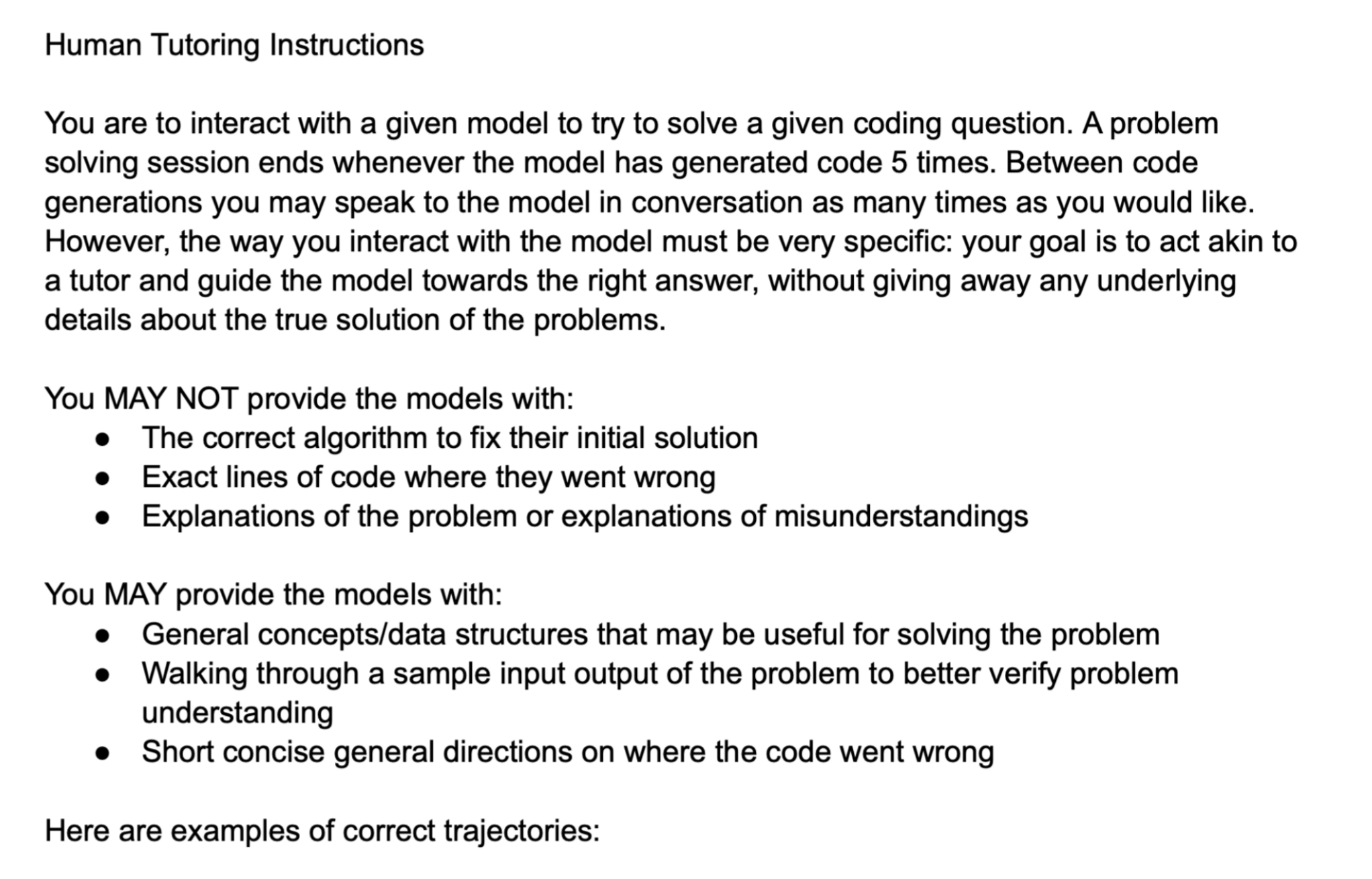} %
  \caption{The prompt given to human tutors when interacting with the model. We manually inspect trajectories to ensure rules were followed. To limit the amount of conversation, we only allow the model to generate code 5 times: however, the code is only allowed to be executed once to simulate a pass@1 setup.}
  \label{fig:HumanTutoringPrompt}
\end{figure}
\subsection{Human-in-the-loop sample trajectory}
\begin{figure}[htbp]
  \centering
  \includegraphics[width=0.6\textwidth]{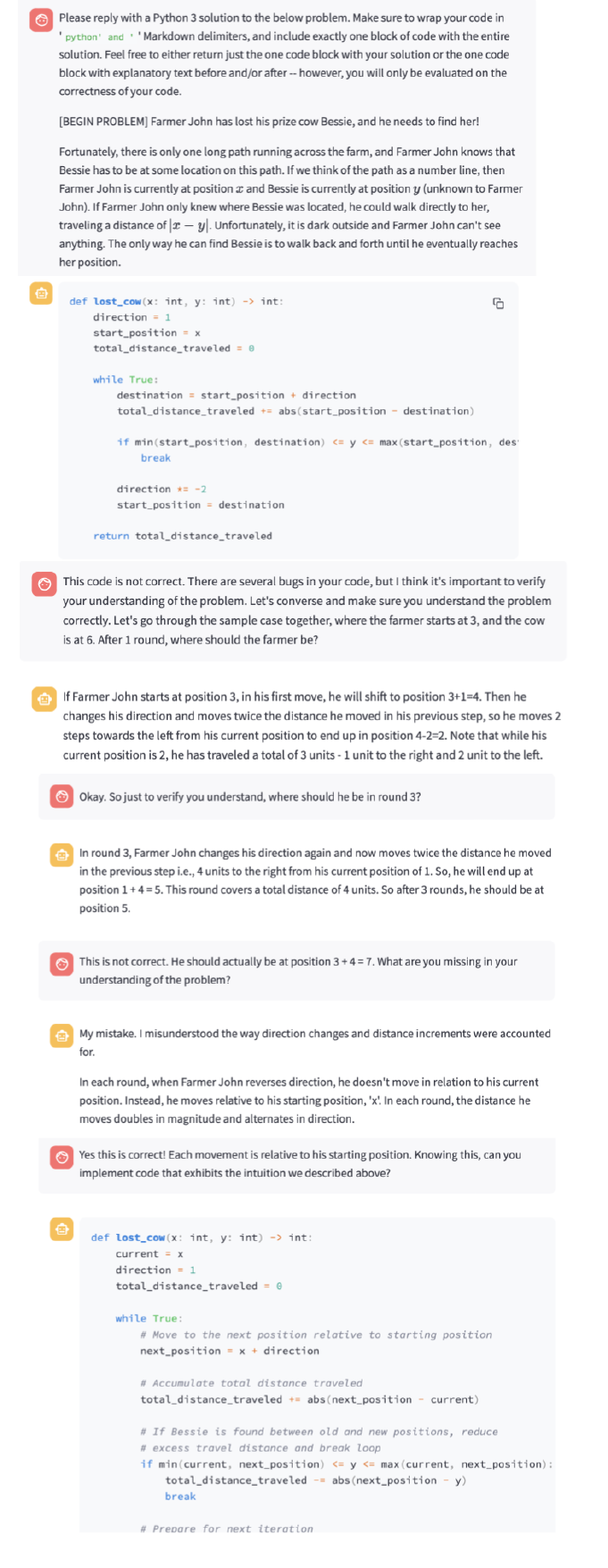} %
  \caption{Sample human tutoring trajectory on the problem 786\_bronze\_the\_lost\_cow}
  \label{fig:SampleTrajectory}
\end{figure}
\label{human_sample_trajectories}
\end{document}